\documentclass[acmlarge]{acmart}
\usepackage{subfigure}

\AtBeginDocument{%
  \providecommand\BibTeX{{%
    \normalfont B\kern-0.5em{\scshape i\kern-0.25em b}\kern-0.8em\TeX}}}

\setcopyright{acmcopyright}
\copyrightyear{2023}
\acmYear{2023}

\begin{document}

\title{A Learnheuristic Approach to A Constrained Multi-Objective Portfolio Optimisation Problem}

\author{Sonia Bullah}
\email{2107762@students.wits.ac.za}
\affiliation{%
  \institution{Computer Science and Applied Mathematics, University of the Witwatersrand}
  \city{Johannesburg}
  \country{South Africa}
}

\author{Terence L. van Zyl}
\affiliation{%
  \institution{Institute for Intelligent Systems, University of Johannesburg}
  \city{Johannesburg}
  \country{South Africa}}
\email{tvanzyl@gmail.com}


\begin{abstract}
Multi-objective portfolio optimisation is a critical problem researched across various fields of study as it achieves the objective of maximising the expected return while minimising the risk of a given portfolio at the same time. However, many studies fail to include realistic constraints in the model, which limits practical trading strategies. This study introduces realistic constraints, such as transaction and holding costs, into an optimisation model. Due to the non-convex nature of this problem, metaheuristic algorithms, such as NSGA-II, R-NSGA-II, NSGA-III and U-NSGA-III, will play a vital role in solving the problem. Furthermore, a learnheuristic approach is taken as surrogate models enhance the metaheuristics employed. These algorithms are then compared to the baseline metaheuristic algorithms, which solve a constrained, multi-objective optimisation problem without using learnheuristics. The results of this study show that, despite taking significantly longer to run to completion, the learnheuristic algorithms outperform the baseline algorithms in terms of hypervolume and rate of convergence. Furthermore, the backtesting results indicate that utilising learnheuristics to generate weights for asset allocation leads to a lower risk percentage, higher expected return and higher Sharpe ratio than backtesting without using learnheuristics. This leads us to conclude that using learnheuristics to solve a constrained, multi-objective portfolio optimisation problem produces superior and preferable results than solving the problem without using learnheuristics.
\end{abstract}





\maketitle

\section{Introduction}
This study explores a multi-objective portfolio optimisation problem by including numerous factors that produce above-average returns while attempting to minimise the investor's overall risk~\cite{paskaramoorthy2020framework}.

This is conducted by introducing realistic constraints into the optimisation model alongside  machine learning, deep learning~\cite{laher2021deep} and metaheuristics to solve non-convex problems. The metaheuristic algorithms explored are Evolutionary Algorithms (EA), namely NSGA-II, R-NSGA-II, NSGA-III  and U-NSGA-III. In contrast, the machine learning models used to enhance these metaheuristic algorithms will be that of the surrogate-assisted variants of each metaheuristic algorithm used~\cite{perumal2020surrogate,stander2020extended}.

In the study by Ma \emph{et al.} \cite{b4}, the researchers compare machine learning and deep learning models with portfolio optimisation models to determine the best combination of investment strategies. Ultimately, it is concluded that the Random Forest Model + Mean-Variance with Forecasting (MVF) Model produces the most efficient and profitable results for practical investments. However, the only issue is that a high turnover results in a huge challenge for these models to overcome. A high turnover implies that the transaction costs or transaction fees (both terms can be used interchangeably) will be quite high. Since these costs are very much present in realistic trading scenarios, they should not be overlooked when analysing the portfolio optimisation problem. Hence, this study will introduce them and other realistic factors as constraints in the portfolio optimisation problem. Adding these additional constraints can be done by employing a hybrid approach, which includes merging metaheuristics with machine learning and deep learning techniques. This approach can be called ‘learnheuristics’ \cite{b5}.

Overall, this study will focus on how realistic constraints, such as transaction costs and holding costs, can be introduced into the portfolio optimisation problem in such a way as to account for their presence in realistic trading scenarios as well as how learnheuristics can be used to assist in solving a constrained, multi-objective and non-convex portfolio optimisation problem. By introducing a unique, hybrid approach, this study will solve the current issue of unrealistic trading scenarios in portfolio optimisation problems while producing superior results compared to those produced by alternative methods and benefiting the investor simultaneously.

\section{Background}

\subsection{Portfolio Optimisation}

Portfolio optimisation forms the basis of this study and is a fundamental problem faced by many investors \cite{b9}. The portfolio optimisation process comprises selecting the optimal portfolio from all possible portfolios to maximise investor wealth while minimising risk. A popular way of doing this includes using the Sharpe ratio, which can be directly optimised \cite{b1}.



One of the major contributions to the portfolio selection problem is Markowitz’s theory of portfolio selection \cite{b8}. Markowitz derives a model that assumes that investors aim to maximise their returns for a particular level of risk, or the converse. Many other studies have extended Markowitz’s theory to include constraints. Kizys \emph{et al.} \cite{b6} acknowledge that the complexity of this model increases significantly when further constraints are added. However, metaheuristics can be used to overcome such obstacles.

Doering \emph{et al.} \cite{b5} extend Streichert \emph{et al.} \cite{b11} by formulating an unconstrained portfolio optimisation problem as per Markowitz's mean-variance theory in such a way that minimises risk as follows:
$$Min \sum_{i=1}^{N} \sum_{j=1}^{N} \omega_i \omega_j \sigma_{ij}$$ 
and maximises profit or returns in the following manner:
$$Max \sum_{i=1}^{N} \omega_i \mu_i$$
subject to: 
$$\sum_{i=1}^{N}  \omega_i = 1  , \; \;\; \; where \; 0 \leq \omega_i \leq 1, \forall i = 1, 2, ... N, $$

and N represents the available assets, $\mu_i$ represents the expected return of asset $i$, $\sigma_{ij}$ represents the covariance between two assets $i$ and $j$, and $\omega_i$ represents the decision variables of the portfolio. This study will adjust such models to include realistic constraints such as transaction costs.

Furthermore, portfolio optimisation can be classified as a multi-objective optimisation problem \cite{b2} with the target of finding solutions that can illustrate the trade-off present for different objectives \cite{b9}. A multi-objective portfolio optimisation problem requires two objectives to be met \cite{b11}.

Machine learning and deep learning models can be combined with traditional optimisation models \cite{b4} to find the best combination for practical investments. These ideas will be explored in the sections below.

\subsection{Machine Learning Models and Deep Learning Models for Multi-Objective Portfolio Optimisation}

Zhang \emph{et al.} \cite{b1} focuses primarily on using deep learning models to evaluate portfolio optimisation problems by first providing an objective function in which the Sharpe ratio is defined and maximised using gradient ascent.

Ma \emph{et al.} \cite{b4} make use of popular machine learning models, such as Support Vector Regression (SVR) and Random Forest (RF), as well as deep learning models, such as Deep Multilayer Perceptron (DMLP), LSTM Neural Network and Convolutional Neural Network (CNN). Furthermore, they pair each model with a traditional portfolio optimisation model, namely the Mean-Variance and Omega Models.

In addition to using machine learning and deep learning models, many studies also introduce hybrid algorithms. This can be seen in Kizys \emph{et al.} \cite{b6}, in which a new algorithm, \emph{ARPO}, is used to consider inevitable, real-world constraints that many researchers refrain from analysing. Similarly, van Zyl \emph{et al.} \cite{b2} implement \emph{ParDen} to include machine learning models that are either generative or discriminative and use them as a surrogate to reduce costs that arise from backtesting procedures.

Optimisation problems depend on time-intensive functions due to the problem’s computationally expensive nature \cite{b16}. Implementing such functions influences the computational overhead of the algorithm, which is used to discover new solutions every time the function is called. Therefore, a \emph{surrogate} model can be incorporated to accelerate the convergence and approximate the time-intensive functions of the models and algorithms currently being used to solve these optimisation problems.

Surrogate-assisted optimisation satisfies the objective of producing efficient approximate solution evaluations (ASEs), which are then used to minimise expensive solution evaluations (ESEs), all while providing a minimal approximation error. Overall, the convergence of the optimisation algorithm being utilised will improve \cite{b16}.

\subsection{Transaction Costs And Holding Costs}

Transaction costs arise from turnover; the higher the turnover, the more transaction costs will increase. This suggests that the performance of a model will differ with and without factoring in the relevant transaction costs \cite{b4}. Ma \emph{et al.} \cite{b4} extends \cite{b2,b3,b5,b6} and \cite{b7}, in which the concept of additional costs is discussed briefly and where transaction costs are recognised as a major constraint in the realistic portfolio optimisation problem. 

Likewise, holding costs are also important to consider in practical investment and trading scenarios. The simplest holding cost model accounts for charges incurred for borrowing assets when taking a short position \cite{b10}. However, these models can become significantly more complex under certain conditions. Similar to transaction costs, holding costs generally consist of non-negative values. However, negative values are not unlikely.

Trading costs and holding costs can be combined to formulate a multi-objective optimisation problem with the inclusion of a  leverage constraint as done by van Zyl  \emph{et al.} \cite{b2}. This is shown by:
$$ 
\text{Maximise:}\quad
w'\mu - \frac{\gamma}{2} w' \sum w - \gamma_t \phi^{trade} (w-w_0) - \gamma_h \phi^{hold} (w)
$$ 
$$
\text{Subject to:}\quad
\lVert w \rVert_1 \leq L^{max} 
$$ 
where $ w_0 $ denotes the initial portfolio, $ w - w_0 $ represents the trades needed to secure $ w $, $ \phi^{trade} $ is the trading cost function with the trade-off parameter $ \gamma_t $, $ \phi^{hold} (w)$ is the trading cost function with the trade-off parameter $ \gamma_h $ and, finally, $ L^{max} $ denotes the maximum leverage. 

It can be noted that a fully-invested portfolio, in which a long position is maintained, has a leverage of 1.

\subsection{Introducing Portfolio Constraints}

Most portfolio optimisation problems consider some form of risk management, and many research papers fall short of considering stochastic optimisation problems, which include using constraints in optimisation models \cite{b5}.

Abate \emph{et al.} \cite{b7} consider portfolio constraints, as considering mean-variance optimisation alone leads to unfair asset allocations. They also implement a list of constrained strategies. These strategies include classical constraints, flexible portfolio constraints, norm-based constraints, variance-based constraints, tracking error volatility constraints and beta constraints. They note that tighter constraints can produce diversified portfolios and that choosing the proper constraints plays a vital role in the evaluation process. Investors should typically choose constraints that align with their target or goal. 

Kizys \emph{et al.} \cite{b6} extend the work of Doering \emph{et al.} \cite{b5} and Abate \emph{et al.} \cite{b7}, in which both articles consider constraints that are unavoidable for investors to ignore in realistic investment scenarios.

The transaction costs play a significant role when choosing constraints. Constrained portfolio optimisation models are favourable to investors and can be seen as an effective alternative to classic investment approaches \cite{b5}.

\subsection{Metaheuristics for Solving Optimisation Problems}

Metaheuristic algorithms can provide solutions to portfolio optimisation problems that are non-convex. In general, the single-period optimisation problem (SOP) can be efficiently solved given that the problem is convex \cite{b10}. However, once this is adjusted to become a multi-objective optimisation problem (MOP) with non-convex constraints, the problem becomes NP-hard and is more complicated to solve. Metaheuristics play a vital role in this case as they can be used to solve NP-hard optimisation problems and provide exceptional solutions \cite{b5}, while considering the realistic, non-convex constraints introduced into the problem. 

Doering \emph{et al.} \cite{b5} classify metaheuristics into two groups: nature-inspired and non-nature metaheuristics. This study will primarily focus on a sub-class of nature-inspired metaheuristics, evolutionary algorithms. In particular, it will focus on the Genetic Algorithm (GA). However, this study focuses on multi-objective problems, so multi-objective metaheuristics must be incorporated. Therefore, the various forms of genetic algorithms, namely, NSGA-II, R-NSGA-II, NSGA-III and U-NSGA-III \cite{b2}, will be explored.

NSGA-II is a genetic algorithm based on non-domination and is popular for solving multi-objective optimisation problems \cite{b13}. First, the algorithm initialises the population and considers any constraints that may be specified. Next, the population goes through a non-dominated sort. Each individual is then given a rank or fitness value based on the front to which they belong. Furthermore, a crowding distance is assigned to each individual in the population, quantifying how close the relevant individual is to their neighbours. Next, a binary tournament selection process is used to select parents from the population, and offspring are produced from the selected population. The relevant population and offspring are again sorted using a non-dominated sort procedure, and only the best individuals are chosen.

The R-NSGA-II algorithm is an extension of the NSGA-II algorithm with an improved survival selection process \cite{b2}. In this algorithm, a smaller Euclidean distance between parents and the reference point is favoured or preferred \cite{b15}. The algorithm begins the same way as NSGA-II, resulting in a single, merged population. The merged population is then segmented into various non-dominated levels using non-dominated sorting. Then each front of the population is clustered to ensure that the solutions are diverse. The whole population size is then diminished to match that of the parent population \footnote{In the case of not all members being selected, the ones with the smaller Euclidean distances between the member itself and the reference point will be conserved.} \cite{b15}. The difference between this algorithm and NSGA-II is that solutions are based on the rank or fitness value while segmenting the front \cite{b2}.

Unlike NSGA-II, which uses selection based on rank or fitness and crowding distance, NSGA-III uses reference directions to ensure diversity amongst the solutions \cite{b14}. The NSGA-III algorithm begins with a random population and disperse reference points on a hyper-plane and aims to find only one population member per reference point. The population is segregated into various non-dominated levels, and then  non-dominated sorting is performed, just as the NSGA-II algorithm does. After this, an offspring population is produced from which a combined population is formed. Next, points are iteratively selected until a point is reached at which all solutions from a complete front cannot be selected. The final front will not be able to be fully selected. However, only a few solutions from this front need to be selected in this case. Then, each population member and member of the final front is linked to a reference point obtained from the minimum perpendicular distance of each member. Next, a niching strategy is implemented and followed to obtain members of the final front corresponding to the population's marginalised reference points. Population members associated with these points are preferred. The algorithm attempts to find one population member corresponding to each reference point close to the Pareto-optimal front \cite{b14}.

The population will be segmented into non-dominated fronts using a non-dominated sorting process for a multi-objective optimisation problem. U-NSGA-III uses a niched tournament selection process as this has a superior performance compared to random selection \cite{b2}. This places weight on non-dominated solutions over dominated solutions and solutions nearer to the reference directions than those farther away. Thereafter, the algorithm follows the same process as that of the NSGA-II algorithm \cite{b14}.

\subsection{The Learnheuristic Approach}

\emph{‘Learnheuristics’} is a hybrid approach that combines metaheuristics with machine learning or statistical learning models \cite{b5}. This study will use a learnheuristic approach to pair each of the four multi-objective genetic algorithms with a surrogate model. The surrogate-assisted algorithms will then be used to evaluate the portfolio optimisation problem.

Calvet \emph{et al.} \cite{b12} explain that the hybridisation of machine learning and metaheuristics can be split into two classes, namely one in which machine learning is used to enhance metaheuristics and the other class being one where metaheuristics are used to improve the machine learning models or techniques used.

Using machine learning to enhance metaheuristics can be split into specifically-located and global-level hybridisations. Specifically-located hybridisations take on approaches such as fine-tuning parameters, quick initialisation, inexpensive evaluation techniques, population management, introducing the knowledge in operators and, lastly, using machine learning techniques as local searches. 
On the other hand, global hybridisations focus on reducing the search space, algorithm selection, using hyperheuristics (which are search methods used for choosing heuristics that will provide solutions to computational search problems), cooperative strategies and, finally, using new types of metaheuristics. \newline 
Conversely, using metaheuristics to improve machine learning techniques spans topics such as classification, regression, clustering and rule mining.

Taking on a learnheuristic approach implies that either one or both of the classes mentioned above be implemented. Table 1 indicates that while the related work either acknowledges or implements methods that consider some form of realistic constraints, only the survey presented by Doering \emph{et al.} \cite{b5} focuses on metaheuristics to solve multi-objective optimisation problems. Therefore, this study will combine techniques outlined in the related work and attempt to enhance metaheuristics by using machine learning techniques and producing improved results.

\begin{table}
  \caption{A summary of the work related to this study.}
  \label{tab:1}
  \begin{tabular}{p{2cm}p{2cm}p{2cm}p{3cm}p{6cm}}
    \toprule
    Author & Optimisation & Constraints & Algorithms & Advantages\\
    \midrule
       Ma \emph{et al.} &  Single- \& Multi-objective & With constraints & Machine learning \& Deep learning & Provides solutions to NP-hard optimisation problems; produces high-quality solutions.\\
       Zhang \emph{et al.} &  Single-objective &  With constraints & Deep learning & Improved performance; method is rational and practical.\\
       Doering \emph{et al.} & Single-objective &  With and without constraints & Metaheuristic & Yields a high turnover.\\
  \bottomrule
\end{tabular}
\end{table}

\section{Methodology}

To test whether or not a learnheuristic approach to a multi-objective portfolio optimisation problem will produce higher returns while minimising the risk of an investment, we generate a hybrid learnheuristic algorithm and compare the results to those of the baseline algorithms. In the past, research surrounding multi-objective portfolio optimisation, such as the research conducted by Ma \emph{et al.} \cite{b4}, has been explored without realistic constraints, such as transaction costs. Therefore, while attempting to maximise returns and minimise risk concurrently, our study will factor in these constraints in a learnheuristic approach similar to the process outlined in the study by Calvet \emph{et al.} \cite{b12}.

\begin{figure}[htb!]
    \centering
    \includegraphics[width= 8cm]{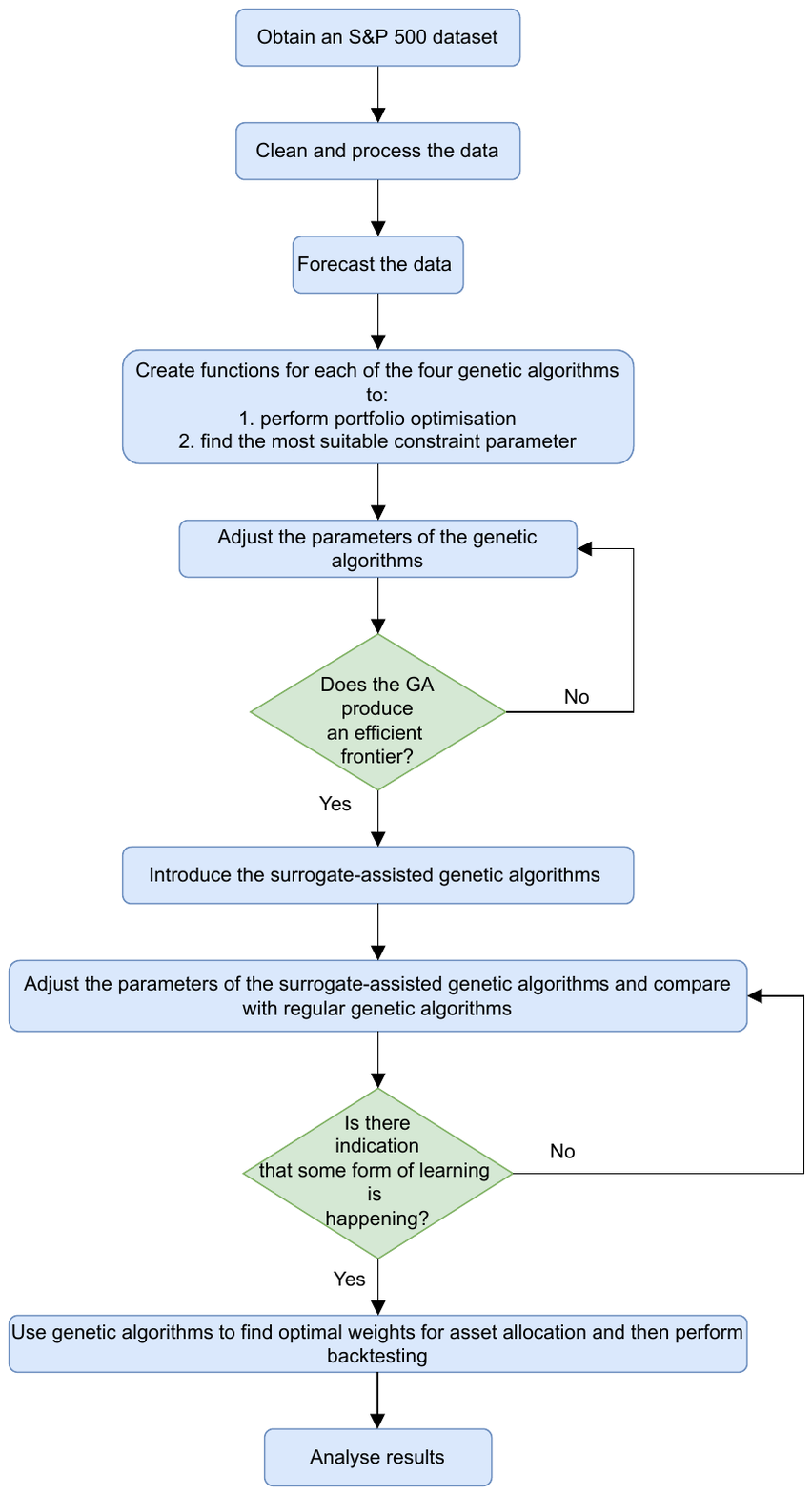}
    \caption{A summary of the methodology followed in this study.} 
    \label{fig:1}
\end{figure}

The implementation of this study can be separated into five main interdependent sections. The sections will all be described in detail below and are summarised in Fig. \ref{fig:1}.

\subsection{Data Cleaning And Exploration}

The S\&P 500 data was pre-processed to ensure correct datatypes and eliminate redundant attributes. Duplicates, missing values and null values were also dropped. We used only the top ten stocks regarding expected returns as our portfolio to conduct the rest of the study.

\subsection{Forecasting}

To forecast the data, some form of noise needs to be added. This noise represents an error that will increase or decrease predictions. This makes it difficult to make accurate predictions about the data. We use a mean of 0 and a standard deviation of 0.1 to acquire random samples from a normal distribution. This, in turn, outputs an array containing the noise, which is now added to the data, resulting in forecasted data values. Lastly, any null values need to be dropped before the portfolio optimisation.

\subsection{Genetic Algorithms}

In this study, the four genetic algorithms are needed to satisfy two main objectives, namely, optimising the portfolio and then finding the optimal constraint value simultaneously and, secondly, producing the best set of weights that can be used to perform backtesting. All genetic algorithms are implemented using the \emph{Pymoo} \cite{b18} library, and the portfolio optimisation method followed is dependent on \emph{PyPortfolioOpt} \cite{b17}.

First, an objective function is adjusted to perform portfolio optimisation on the data. This is done by obtaining the forecasted dataset's expected returns and sample covariance. The expected returns are computed using the Capital Asset Pricing Model
$$ R_i = R_f + \beta_i (E(R_m) - R_f), $$
where $R_f$ represents the risk-free rate, $\beta_i$ is the beta (or sensitivity) of asset \emph{i} and $(E(R_m) - R_f)$ the market risk premium.
In addition, the sample covariance is obtained using the Ledoit-Wolf covariance shrinkage estimator 
$$ \hat{\Sigma} = \alpha F + (1 - \alpha)S, $$ 
where $\alpha$ represents the shrinkage constant, $F$ the a structured estimator and $S$ a sample covariance matrix.

This, in turn, allows us to find optimal portfolios and produce an efficient frontier once the results are correctly plotted. Furthermore, it should be noted that this method will solve an unconstrained portfolio. However, the focus of this study is on constrained portfolios. As such, we add a constraint to the optimisation problem.

Once an efficient frontier has been produced, we proceed by adding a constraint to the efficient frontier. Instead of adding an arbitrary constraint with a random value, we search for the best possible parameter. Once this has been found, the Sharpe ratio is maximised, and the overall portfolio performance is returned.

The genetic algorithms are executed by randomly generating an initial population and assigning a fitness score to each individual. Next, individuals go through the selection and reproduction stages before crossover and mutation occur, in which new genetic information is added to the child population. Finally, generational replacement occurs, and then the algorithm terminates based on a stopping condition.

Once each genetic algorithm returns the array of portfolio performance results obtained, we adjust the array by removing the Sharpe Ratio values from each sub-array as we do not want a three-dimensional plot. This leaves us with sub-arrays of size two, which contain the optimal portfolio's return and volatility (or risk). These results are then plotted.

This process is run ten times on a population of 12 over 30 generations. We compare results by plotting all four genetic algorithms on the same axis. To compare the success of each algorithm, we note the average time taken for each algorithm to run to completion and the average hypervolume produced by the algorithm. Firstly, we begin a timer each time an algorithm starts and end it once the algorithm completes its last generation. The algorithms' average time (in seconds) is returned at the end. After each algorithm runs, we use the hypervolume function from the \emph{Pymoo} \cite{b18} library. A hypervolume is a performance indicator which uses a reference point to calculate the area influenced by the solution set concerning a specific reference point \cite{b18}. It should be noted that higher hypervolumes suggest better-performing algorithms. Once we have the relevant hypervolumes for each generation of each algorithm, we calculate the average hypervolume from all ten runs and compare the algorithms to each other to determine how well they perform over the total number of generations.

Within the objective function, we also add an objective to the efficient frontier, which aims to adjust the gamma value. This parameter for L2 regularisation can be increased when requiring further non-negligible weights. The genetic algorithms search for the best possible value of gamma, which, in turn, produces the optimal set of weights that are used to perform backtesting. This process is elaborated on in the next section. Like the first case, once the set of weights for each stock is found, the portfolio is optimised, and the maximum Sharpe ratio is found. The final weights of each stock of the optimised portfolio are then added to an array, which is utilised in the backtesting procedure.

\subsection{Learnheuristics: Surrogate-Assisted Genetic Algorithms}
Once our genetic algorithms run, we adjust each function to include a surrogate model. We call the GPSAF implementation from the \emph{Pysamoo} \cite{b19} framework to do this. Furthermore, we manually fine-tune the \emph{alpha}, \emph{beta}, \emph{n\_max\_doe}, \emph{n\_max\_infills}, \emph{n\_offsprings} and \emph{seed} parameters to produce an optimal convergence rate for each algorithm. We continue adjusting these parameters until the surrogate-assisted genetic algorithms are `learning' throughout the iterations. We eventually produce an improved solution compared to the non-surrogate-assisted genetic algorithms. Again, these learnheuristic algorithms are run ten times on a population size of 12 over 30 generations, with the average time taken and average hypervolume being recorded at the end. We compare results by plotting each genetic algorithm alongside the surrogate-assisted version of the algorithm on the same axis and take note of the difference in convergence rate, average hypervolume and the average time taken for the algorithms to run to completion.

\subsection{Backtesting}

We run simulations or backtests to help optimise our strategy. To do this, we call the \emph{Backtest} function from the \emph{portfolio-backtest} \footnote{The documentation can be found on: https://pypi.org/project/portfolio-backtest/} library. The weights obtained by the genetic algorithm are substituted into the function and the target return, target variance, start period and end period. We first run a backtest using the function's built-in method of determining the weights for each asset and then use each of the four surrogate-assisted genetic algorithms to search for the optimal set of weights that can then be passed into the backtesting function. Lastly, we record the data, namely the annual volatility, expected annual return and Sharpe ratio outputted from each backtest and compare the results.

All algorithms and implementations were run on a Quad-Core Intel Core i5 running macOS 12.3. The complete code, datasets and results can be found on the \href{https://github.com/soniabullah1/ResearchProject2022}{\color{blue}GitHub} repository.

\section{Results and Discussion}

In this section, we present the results of the four multi-objective genetic algorithms used to optimise the portfolio and then consider the limitations and future work. 

\subsection{Results}

Based on the results presented in Fig. \ref{tab:2} below, it is evident that the NSGA-III algorithm runs to completion in the shortest amount of time, followed by the U-NSGA-III algorithm and then the R-NSGA-II algorithm, while the NSGA-II algorithm takes the longest time to run. However, given that the NSGA-II algorithm produces the highest hypervolume followed by the R-NSGA-II, U-NSGA-III, and, lastly, NSGA-III  algorithms, it suggests that the best performing algorithm is NSGA-II.

\begin{figure}[!htb]
\centering

\subtable[The average hypervolume and average time taken (in seconds) for each algorithm to run over 30 generations with a population size of 12.]{%
    \raisebox{50pt}{%
\resizebox{0.48\textwidth}{!}{%
    \begin{tabular}{l|ccrr}
      \toprule
    Algorithm & Generations & Population & HV & Time (sec)\\
    \midrule
       NSGA-II & 30 & 12 & 0.7225 & 1.0115\\
       R-NSGA-II & 30 & 12 & 0.5790 & 0.7941\\
       NSGA-III & 30 & 12 & 0.4986 & 0.7557\\
       U-NSGA-III & 30 & 12 & 0.5126 & 0.7681\\
  \bottomrule
    \end{tabular}
    }
    }%
    \label{tab:2}
  }%
\hfill
\subfigure[A comparison of all four genetic algorithms with a population size of 12 over 30 generations.]{%
\includegraphics[width=0.48\linewidth]{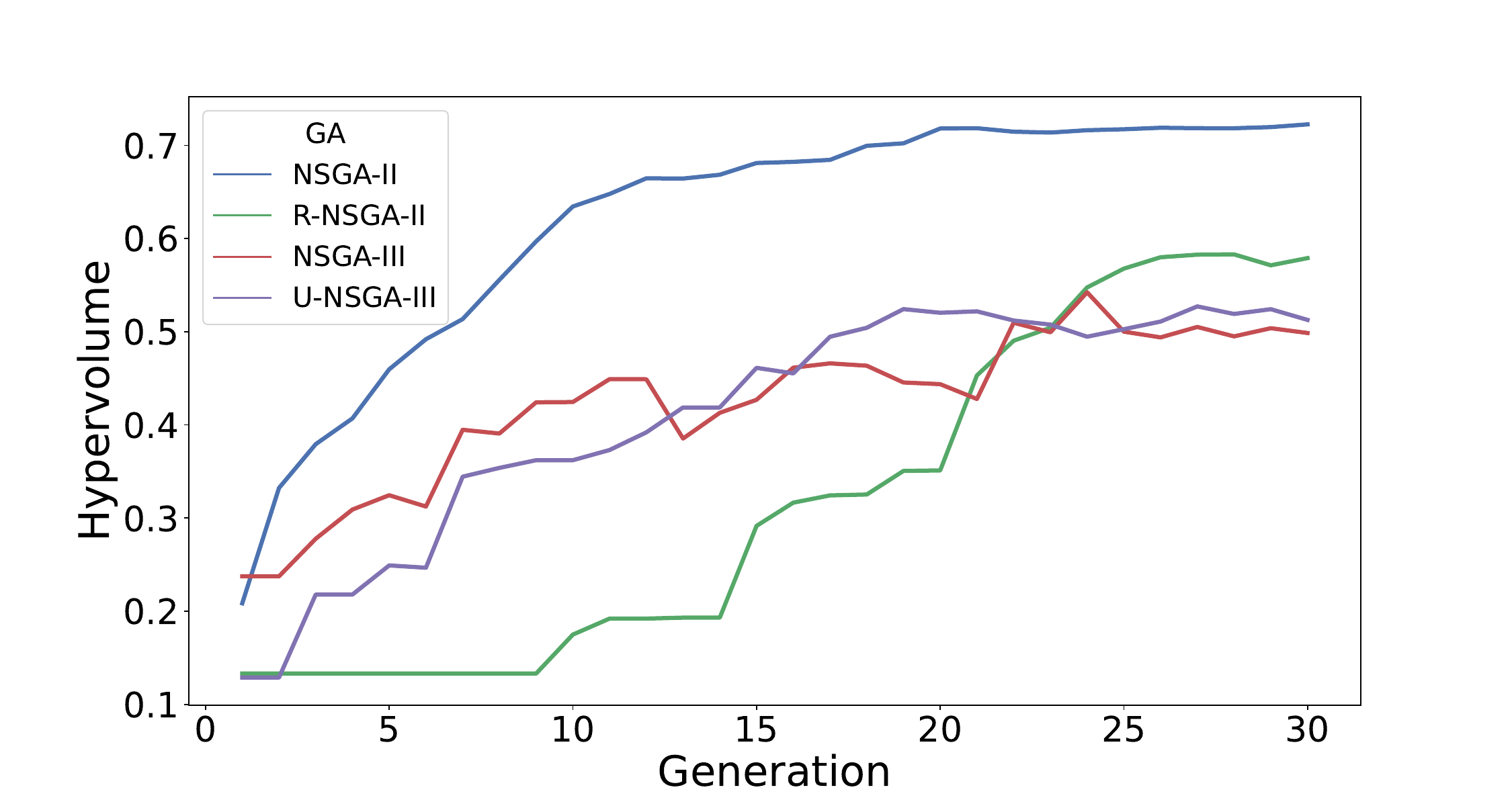}
\label{fig:2}
}

\caption{A comparison of the different genetic algorithms.}
\label{fig:comparison}
\end{figure}


Based on Fig. \ref{fig:2}, it can be seen that the NSGA-II algorithm is the first to converge and produces a smoother line plot. The remaining three algorithms contain more spikes in their respective line plots, however, the general outcome suggests that the hypervolume increases as the number of generations increases.


Next, we present the results of each of the four learnheuristic algorithms (or surrogate-assisted genetic algorithms). Fig. \ref{tab:3} indicates that the surrogate-assisted R-NSGA-III algorithm runs to completion in the shortest amount of time, followed by the U-NSGA-III algorithm and then the NSGA-III algorithm, while the NSGA-II algorithm takes the longest time to run. However, similar to the non-surrogate variant, the NSGA-II algorithm yields the highest hypervolume, thus making it the most efficient algorithm.

\begin{figure}[!htb]
\centering

\subtable[The average hypervolume and average time taken (in seconds) for each surrogate-assisted metaheuristic algorithm to run over 30 generations with a population size of 12.]{%
    \raisebox{5pt}{%
\resizebox{0.48\linewidth}{!}{%
    \begin{tabular}{l|ccrr}
      \toprule
    Algorithm & Generations & Population & HV & Time (sec)\\
    \midrule       
       SA\_NSGA-II & 30 & 12 & 0.7543 & 18.0166\\
       SA\_R-NSGA-II & 30 & 12 & 0.7181 & 15.2820\\
       SA\_NSGA-III & 30 & 12 & 0.6403 & 16.6078\\
       SA\_U-NSGA-III & 30 & 12 & 0.7093 & 16.4171\\
  \bottomrule
    \end{tabular}
    }
    }%
    \label{tab:3}
  }%
\hfill
\subtable[The annual volatility, expected annual return and Sharpe ratio (SR) of a tangency portfolio generated from backtests as well as the time taken (in seconds) for each algorithm to run.]{
\resizebox{0.48\linewidth}{!}{%
    \begin{tabular}{l|ccrr}
      \toprule
    Algorithm & Volatility (\%) & Return (\%) & SR & Time (sec)\\
    \midrule
       No GA & 22.4 & 26.6 & 1.10 & - \\
       SA\_NSGA-II & 18.2 & 55.6 & 2.94 & 40.7051 \\
       SA\_R-NSGA-II & 18.2 & 55.6 & 2.94 & 43.0664 \\
       SA\_NSGA-III & 18.2 & 55.6 & 2.94 & 29.5306 \\
       SA\_U-NSGA-III & 18.2 & 55.6 & 2.94 & 28.7683 \\
  \bottomrule
    \end{tabular}
    }
    \label{tab:4}
  }
\caption{A comparison of the different surrogate-assisted metaheuristic algorithms.}
\label{fig:table_comparison}
\end{figure}

\begin{figure}[!htb]
\centering

\subfigure[A comparison of the NSGA-II and surrogate-assisted NSGA-II algorithms.]{%
\includegraphics[width=0.48\linewidth]{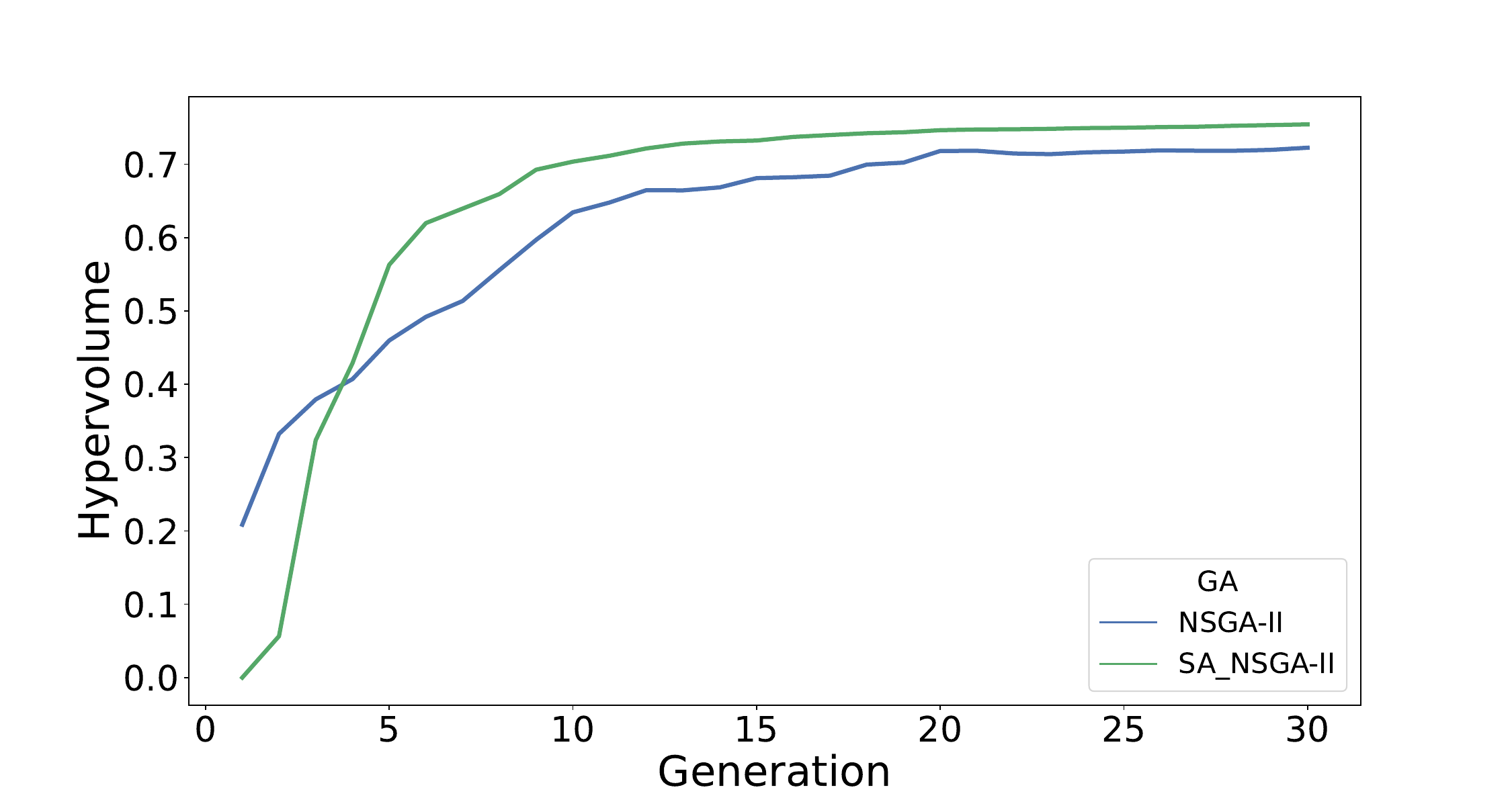}
\label{fig:3}
}
\hfill
\subfigure[A comparison of the R-NSGA-II and surrogate-assisted R-NSGA-II algorithms.]{%
\includegraphics[width=0.48\linewidth]{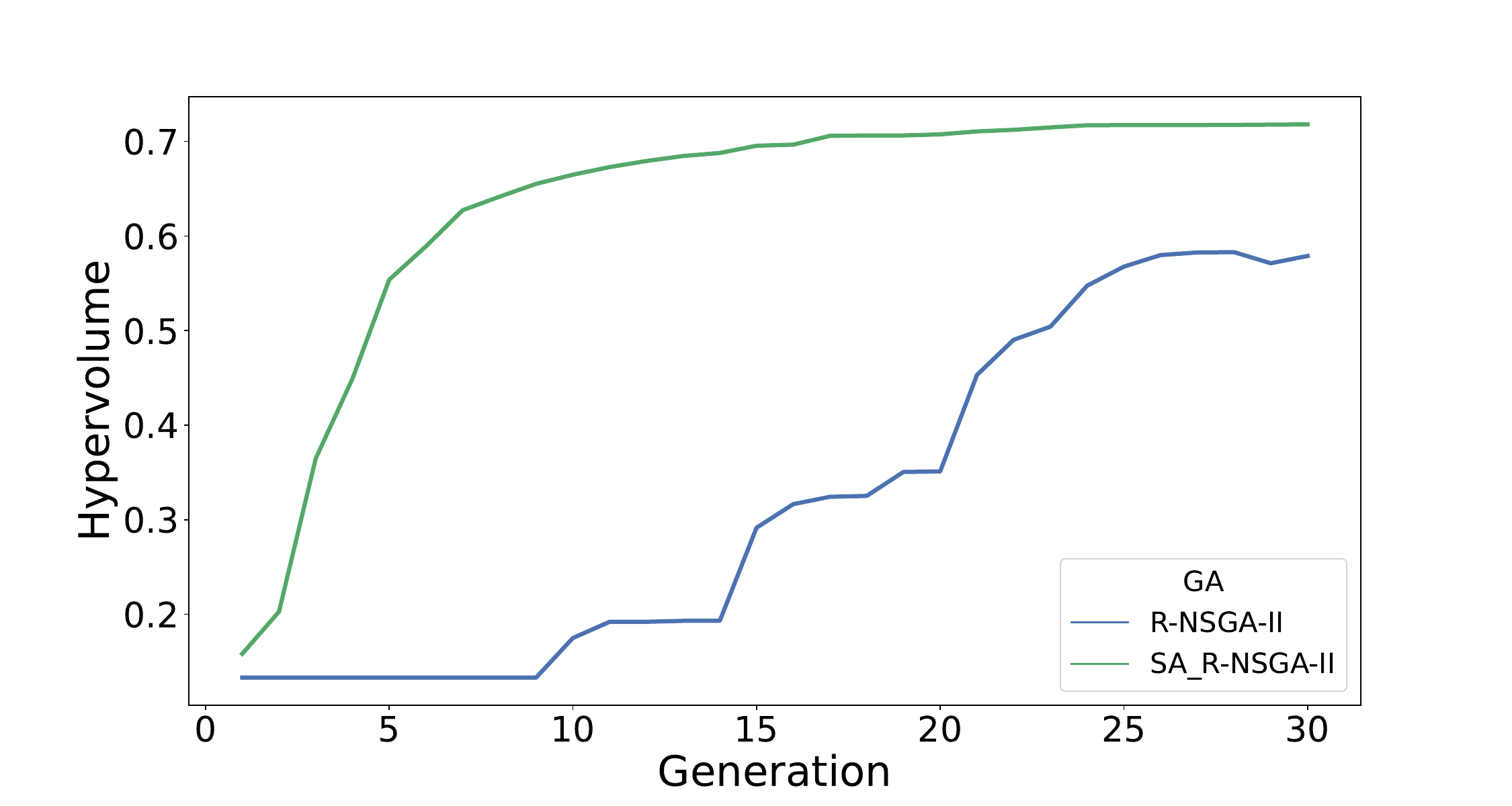}
\label{fig:4}
}

\subfigure[A comparison of the NSGA-III and surrogate-assisted NSGA-III algorithms.]{%
\includegraphics[width=0.48\linewidth]{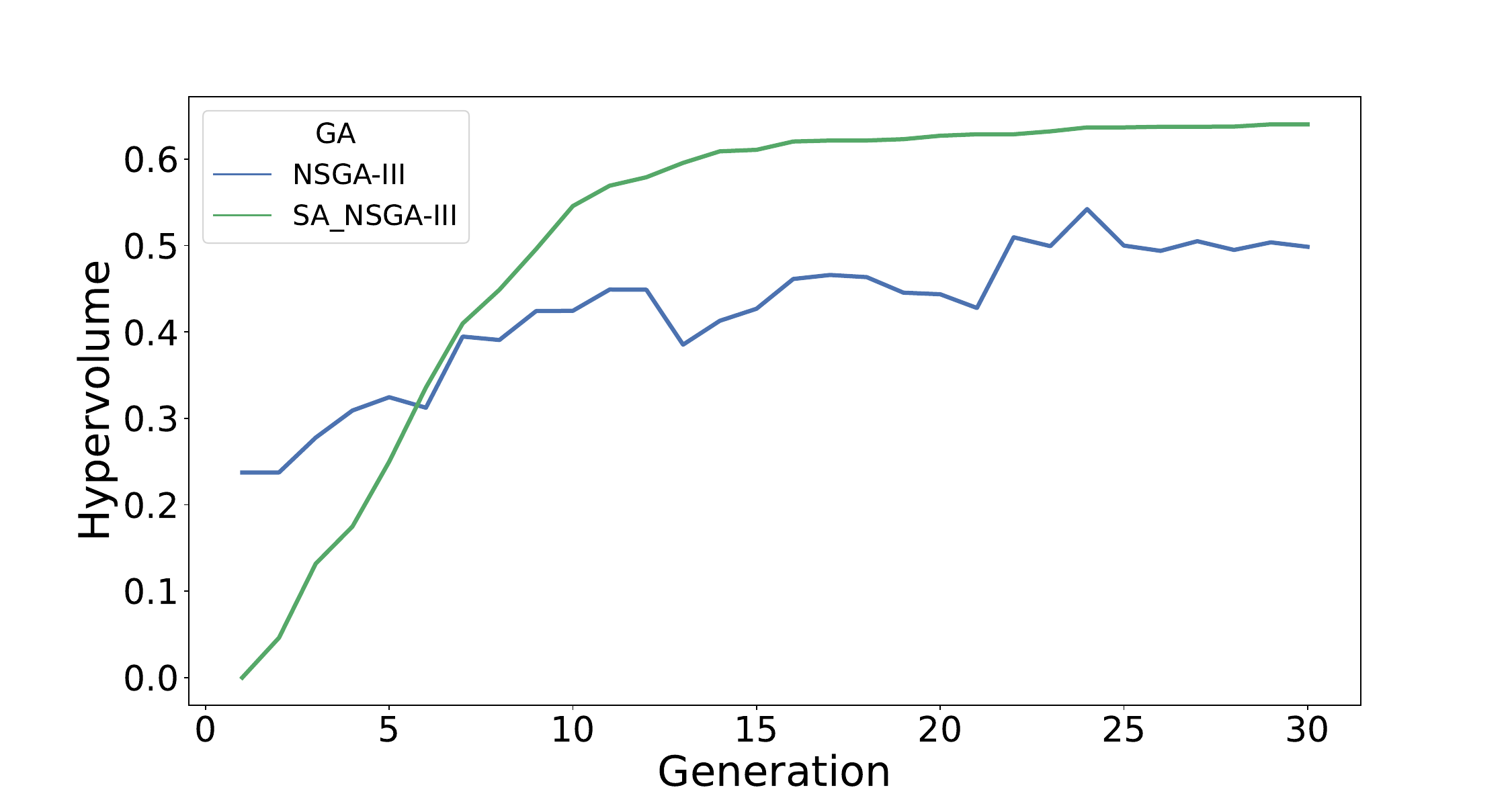}
\label{fig:5}
}
\hfill
\subfigure[A comparison of the U-NSGA-III and surrogate-assisted U-NSGA-III algorithms.]{%
\includegraphics[width=0.48\linewidth]{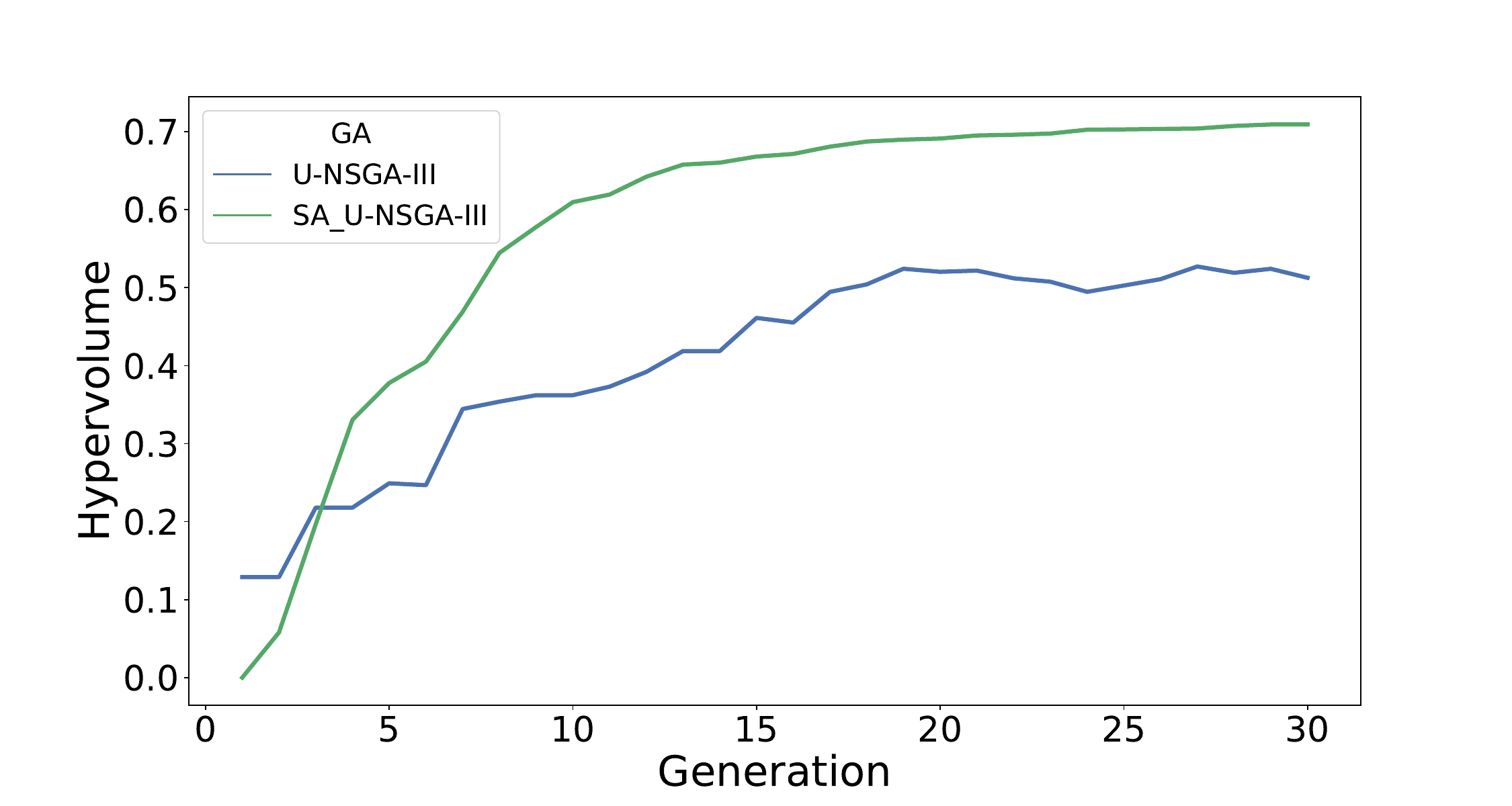}
\label{fig:6}
}

\caption{A comparison of the different algorithms.}
\label{fig:comparison}
\end{figure}

Fig. \ref{fig:3}, Fig. \ref{fig:4}, Fig. \ref{fig:5} and Fig. \ref{fig:6} below each indicate that the learnheuristic algorithms produce higher hypervolumes than the baseline algorithms. By comparing Fig. \ref{tab:2} and Fig. \ref{tab:3}, we observe that while the learnheuristic algorithms take significantly longer to run to completion, they produce much higher hypervolumes and converge faster while also creating smoother plots. As Blank \emph{et al.} \cite{b17} mentioned, the learnheuristic algorithms accelerate the convergence of the multi-objective metaheuristic algorithms used to solve the optimisation problem.

Lastly, we run simulations or backtests on our data and record the annual volatility, expected annual return and Sharpe ratio generated for a tangency portfolio of ten assets. Based on Fig. \ref{tab:4}, we see that the tangency portfolio for each of the four learnheuristic algorithms outperforms the backtest run without a learnheuristic algorithm. The portfolios that run by utilising the surrogate-assisted genetic algorithms to find the optimal weights for asset allocation yield higher returns for a lower risk percentage and produce greater Sharpe ratios. In these cases, the expected annual returns and Sharpe ratios are more than double compared to the backtest run without using a genetic algorithm to find the optimal asset weights. In the study by Zhang \emph{et al.} \cite{b1}, the Sharpe ratio is maximised within the objective function before deep learning models evaluate the portfolio optimisation problem. Similarly, in this study, we solve the optimisation problem by maximising the Sharpe ratio in the objective function, which is then called by each metaheuristic algorithm.

Although each of the four learnheuristic algorithms produces optimal results, Fig. \ref{tab:4} indicates that the surrogate-assisted U-NSGA-III algorithm runs to completion in the shortest time, followed by NSGA-III, NSGA-II and, finally, R-NSGA-II. Therefore, the surrogate-assisted U-NSGA-III algorithm is the most efficient learnheuristic for backtesting purposes. Overall, it is evident that the learnheuristic algorithms produce superior results compared to solving the portfolio optimisation problem without a learnheuristic approach.



\subsection{Limitations}

Despite the findings above, it should be noted that the learnheuristic algorithms are sensitive to specific parameters, such as the \emph{n\_offsprings}, \emph{alpha}, \emph{beta}, \emph{n\_max\_doe}, \emph{n\_max\_infills} and \emph{seed} parameters. If one of these parameters is slightly off, the algorithms may prematurely terminate. Furthermore, the algorithms fail to consider large portfolios as increasing the number of stocks in the portfolio amplifies parameter sensitivity. Hence this study only considers the results produced from a portfolio of ten assets. 

\subsection{Future Work}

\subsubsection{Implementing Search Algorithms}

Since the parameters of the algorithms explored are pretty sensitive, search algorithms, such as a grid search or a random search, can be utilised to find the most appropriate parameters to pass into the algorithms. 

\subsubsection{Increasing The Size of The Portfolio}
Once stable parameters have been obtained, future research can analyse much larger portfolios, such as the complete S\&P 500 index, instead of only the top ten considered in the current study.

\subsubsection{Applying Alternative Multi-Objective Metaheuristics}

Future research can be extended to the current study by implementing different multi-objective metaheuristic algorithms to solve the same problem identified in this study. Possible algorithms to consider include Particle Swarm Optimisation (PSO), Differential Evolution (DE) or Evolutionary Strategies (ES).

\section{Conclusion}

Portfolio optimisation problems must be solved while considering realistic constraints inevitable in practical trading and investment scenarios. This study introduces constraints, such as transaction costs, into a multi-objective portfolio optimisation problem using taking on a hybrid approach termed learnheuristics, which uses machine learning techniques, such as surrogate models, to enhance the metaheuristic algorithms being used to solve an optimisation problem, which is non-convex.

After implementing each of the four learnheuristic algorithms, it is evident that the NSGA-II learnheuristic algorithm is the best-performing algorithm as it reaches a higher hypervolume than the other three learnheuristic algorithms implemented. Furthermore, the hypervolume of each learnheuristic algorithm surpasses the hypervolume of the equivalent non-surrogate variant, which is our baseline algorithm, despite taking much longer for the learnheuristic algorithms to run to completion. 

Based on the findings, we can conclude that the learnheuristic algorithms outperform the baseline algorithms regarding hypervolume and convergence rate. However, they take much longer to run to completion. Overall, using learnheuristics to obtain the weights of each asset for backtesting purposes results in a lower annual risk, higher annual expected return and higher Sharpe ratio compared to backtesting without using learnheuristics. Therefore, it can be concluded that learnheuristics produce superior and preferable results when solving a constrained, multi-objective portfolio optimisation problem.

This study also consists of a few limitations. Firstly, the learnheuristic algorithms are sensitive to many input parameters, and since we aim to reach optimal solutions, finding accurate parameters is vital. Secondly, the implementation of the study is only run on a portfolio consisting of ten stocks, as adding more stocks to the portfolio does not result in optimal solutions due to the sensitivity of the learnheuristic parameters. Therefore, in future research, implementing a search algorithm, such as a random or grid search, to tune the parameters would be beneficial. This, in turn, may also assist with allowing the learnheuristic algorithms to consider portfolios with more than ten assets and still produce optimal results. Furthermore, alternative multi-objective metaheuristic algorithms can also be considered in future research.

\bibliographystyle{ACM-Reference-Format}
\bibliography{sample-base}

\appendix

\end{document}